\documentclass[conference]{IEEEtran}
\IEEEoverridecommandlockouts
\ifCLASSOPTIONcompsoc
  \usepackage[nocompress]{cite}
\else
  \usepackage{cite}
\fi

\ifCLASSINFOpdf

\else

\fi

\usepackage{comment}

\usepackage{bm}
\usepackage{amsmath}
\usepackage{dsfont}
\usepackage{graphicx}
\usepackage{multirow}
\usepackage{booktabs}
\usepackage{subcaption}
\usepackage[dvipsnames]{xcolor}
\definecolor{royalblue}{RGB}{65,105,225}
\usepackage[pagebackref,breaklinks=true,colorlinks,urlcolor=magenta,citecolor=royalblue,bookmarks=false]{hyperref}

\usepackage{listings}
\usepackage{algorithmic}
\usepackage{algorithm}

\begin{document}

\title{Seed Feature Maps-based CNN Models for LEO Satellite Remote Sensing Services}
%\title{A Succinct CNN Model based on Generated Features for LEO Satellite Remote Sensing Services}

%\title{Efficient CNN Deployment on Low-Earth Orbit Satellites for Remote Sensing Services}

\author{
% \IEEEauthorblockN{Page limit: 7-10 pages + references
% Deadline: Feb 12, 2023. \\
% Submission: \url{https://easychair.org/conferences/?conf=ieeeicws2023}}

\IEEEauthorblockA{Zhichao Lu$^1$, Chuntao Ding$^{2*}$\thanks{$^*$Corresponding author}, Shangguang Wang$^3$, Ran Cheng$^4$, Felix Juefei-Xu$^5$, and Vishnu Naresh Boddeti$^6$\\
$^1$School of Software Engineering, Sun Yat-sen University, Zhuhai, China.\\
$^2$School of Computer and Information Technology, Beijing Jiaotong University, Beijing, China. \\
$^3$Key Laboratory of Networking and Switching Technology, Beijing University of Posts and Telecommunications, Beijing, China.\\
$^4$Department of Computer Science and Engineering, Southern University of Science and Technology, Shenzhen, China\\
$^5$New York University, New York, NY, USA. \\
$^6$Department of Computer Science and Engineering, Michigan State University, East Lansing, MI, USA\\
Email: luzhichaocn@gmail.com, chuntaoding@163.com, sgwang@bupt.edu.cn, ranchengcn@gmail.com,\\ juefei.xu@nyu.edu, vishnu@msu.edu}
}

\maketitle

\begin{abstract}
Deploying high-performance convolutional neural network (CNN) models on low-earth orbit (LEO) satellites for rapid remote sensing image processing has attracted significant interest from industry and academia. However, the limited resources available on LEO satellites contrast with the demands of resource-intensive CNN models, necessitating the adoption of ground-station server assistance for training and updating these models. Existing approaches often require large floating-point operations (FLOPs) and substantial model parameter transmissions, presenting considerable challenges. To address these issues, this paper introduces a ground-station server-assisted framework. With the proposed framework, each layer of the CNN model contains only one learnable feature map (called the seed feature map) from which other feature maps are generated based on specific rules. The hyperparameters of these rules are randomly generated instead of being trained, thus enabling the generation of multiple feature maps from the seed feature map and significantly reducing FLOPs. Furthermore, since the random hyperparameters can be saved using a few random seeds, the ground station server assistance can be facilitated in updating the CNN model deployed on the LEO satellite. Experimental results on the ISPRS Vaihingen, ISPRS Potsdam, UAVid, and LoveDA datasets for semantic segmentation services demonstrate that the proposed framework outperforms existing state-of-the-art approaches. In particular, the SineFM-based model achieves a higher mIoU than the UNetFormer on the UAVid dataset, with 3.3$\times$ fewer parameters and 2.2$\times$ fewer FLOPs.
\end{abstract}


% \fi
\IEEEpeerreviewmaketitle

\section{Introduction}%Late fan-out PolyCNN for remote sensing 

Euroconsult estimates that around 18,500 small satellites will be launched between 2022 and 2031\footnote{https://mundogeo.com/en/2022/08/09/smallsats-18500-satellites-weighing-up-to-500-kg-will-be-launched-between-2022-and-2031/}. 
As of December 2022, SpaceX has launched more than 3,000 low-earth orbits (LEO) satellites, and OneWeb has launched 462 LEO satellites\footnote{https://planet4589.org/space/con/star/stats.html}. 
While primarily serving communication purposes, these satellites also acquire many remote-sensing images. 
The large-scale, multi-spectral, and rich-source characteristics of remote sensing images have attracted significant attention from industry and academia for their inherent advantages in applications such as land-use monitoring~\cite{Caleb@Large}, understanding traffic flow~\cite{Scott@Dynamic}, population migration prediction~\cite{Chen@Nighttime}, and climate change tracking~\cite{Peri@Self}.

\begin{figure}
\centering
\includegraphics[width=0.8\linewidth]{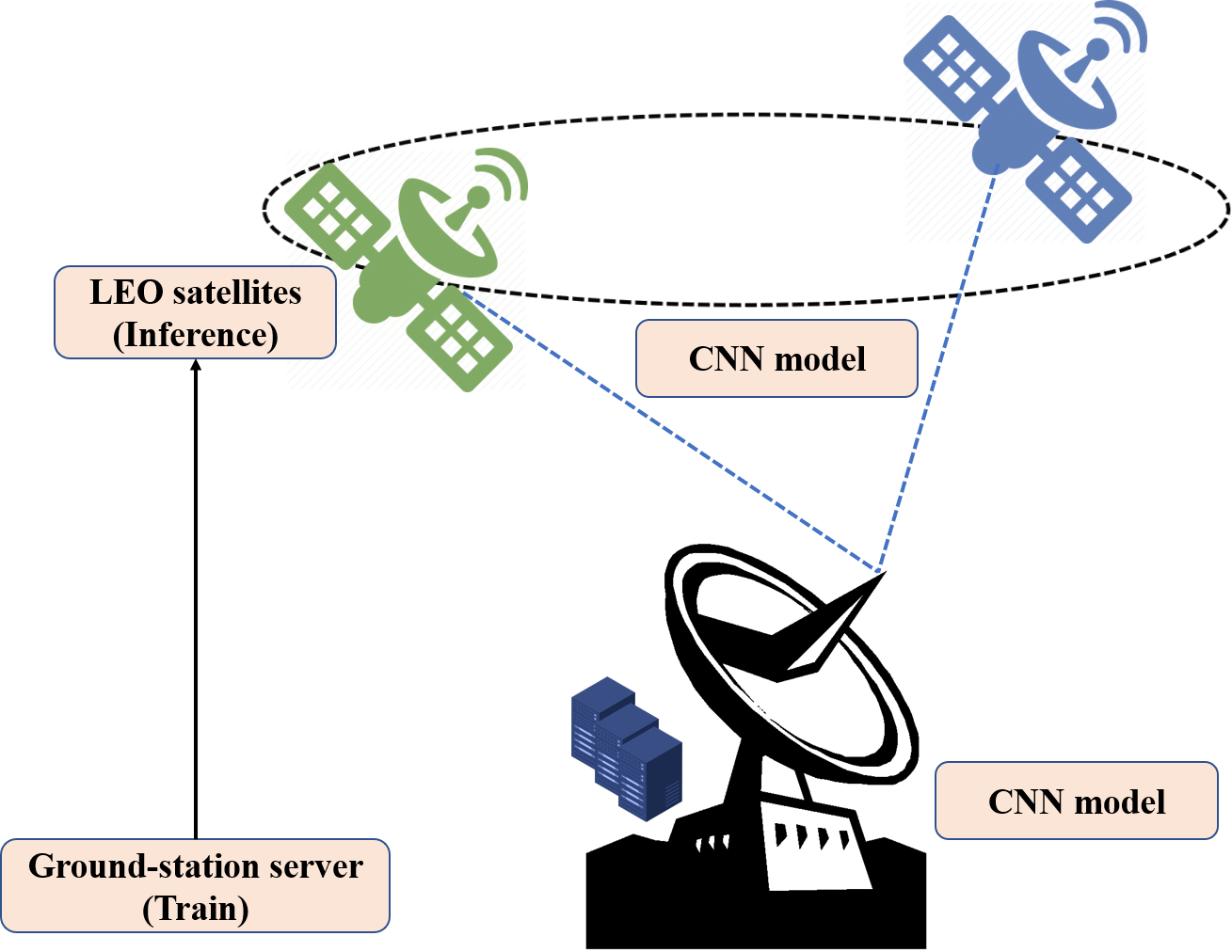}
\caption {System architecture for ground station server-assisted training and updating of CNN models.}
\label{fig:problem}
\vspace{-0.5cm}
\end{figure}

The conventional method for processing remote sensing image data involves satellites capturing the data, sending it to ground station servers, and using convolutional neural network (CNN) models (e.g., ResNet~\cite{resnet, Gao@Res2Net}, VGG~\cite{Karen@Very}) for processing to produce high-performance results.
However, this approach has three limitations:
(i) High Bandwidth Requirements: The transmission of large amounts of raw data from LEO satellites to ground station servers consumes considerable bandwidth.
(ii) Slow Response Time: The communication distance between the satellite and the ground station server is long, which can result in extended transmission delays when transmitting large amounts of data. This is further exacerbated by satellites being powered mainly by solar energy, limiting the transmission power and data transmission rate.
(iii) Increased Bit Error Rates: The link channel is susceptible to various interferences, resulting in a relatively high bit error rate of the transmitted data.

Processing remote sensing image data directly on LEO satellites can partially alleviate these limitations, as transmitting processed results instead of raw remote sensing data can significantly reduce bandwidth consumption, transmission delay, and bit error rates.
However, this approach also introduces two new challenges:
(i) Limited LEO satellite resources and the resource-intensive nature of CNN models pose a challenge for deployment. Performing convolutions in CNN models during inference requires a significant amount of computing resources, as the convolution filter must slide across the remote sensing image multiple times, generating many floating-point operations (FLOPs). For instance, ResNet-50 requires 4.1 billion FLOPs to process an image of size $224\times 224$\cite{resnet, han2020ghostnet}. This makes it challenging to deploy high-performance CNN models on resource-limited LEO satellites.
(ii) The long lifespan of LEO satellites, typically exceeding five years, and the rapid development of CNN models pose challenges in terms of limited bandwidth resources and many model parameters. Therefore, to maintain high service quality, it is necessary to update the CNN model frequently. However, high-performance CNN models often contain a large number of parameters, making the transmission of these parameters during model updates challenging. For example, CoAtNet-7\cite{Dai@CoAtNet} has 2440 million parameters.

To address the above challenges, this paper proposes a ground station server-assisted training and updating the framework for CNN models on LEO satellites processing remote sensing images, as depicted in Fig.~\ref{fig:problem}. 
Our approach begins with introducing SineFM, a novel and efficient feature map generation algorithm. 
SineFM only requires learning one filter parameter, referred to as the seed filter, for each layer in the CNN model. 
By inputting data into the seed filter, a single feature map, referred to as the seed feature map, is generated. 
This seed feature map is then used to generate other feature maps for that layer through nonlinear transformation functions, thereby significantly reducing the FLOPs.

Additionally, we propose that the parameters of the nonlinear transformation function that generates feature maps from the seed feature map be randomly initialized and not trained. 
This allows the parameters to be saved and easily reproducible using a random seed. 
As a result, when updating the CNN model, the ground station server only needs to transmit a small number of seed filters and random seeds to the LEO satellites, significantly reducing the number of transmitted parameters.

Experiments on the ISPRS Vaihingen, ISPRS Potsdam, UAVid, and LoveDA datasets demonstrate the strong performance of our proposed framework for semantic segmentation services. 
Our SineFM-based CNN model slightly improved mIoU  (+0.3\%) while requiring 3.3$\times$ fewer parameters and 2.2$\times$ fewer FLOPs.
In summary, our main contributions are as follows:

\begin{itemize}
\item We propose a novel method, SineFM, to reduce FLOPs in CNN models by using nonlinear transformation to generate feature maps. 
To our knowledge, this is the first work to use this approach. 
The hyperparameters of the nonlinear transformation can be saved using a few random seeds, making it possible for the ground station server to send a small number of parameters for deployment and updating the model on the LEO satellite.

\item Our theoretical analysis of the SineFM-based layer shows that it can approximate the standard convolutional layer well and achieve better performance.

\item Our experiments demonstrate that our proposed framework reduces approximately 3.3 times the model parameter transmission and about 2.2 times the FLOPs.
\end{itemize}

The rest of the paper is structured as follows. First, Section~\ref{RelatedWork} reviews the relevant literature. Then, Section~\ref{Proposed Approach} describes our proposed framework in detail. Then, Section~\ref{Evaluation} presents our evaluation results. Finally, we conclude the paper in Section~\ref{Conclusions}.
\begin {figure*}[t]
\centering
\includegraphics[width=0.85\linewidth]{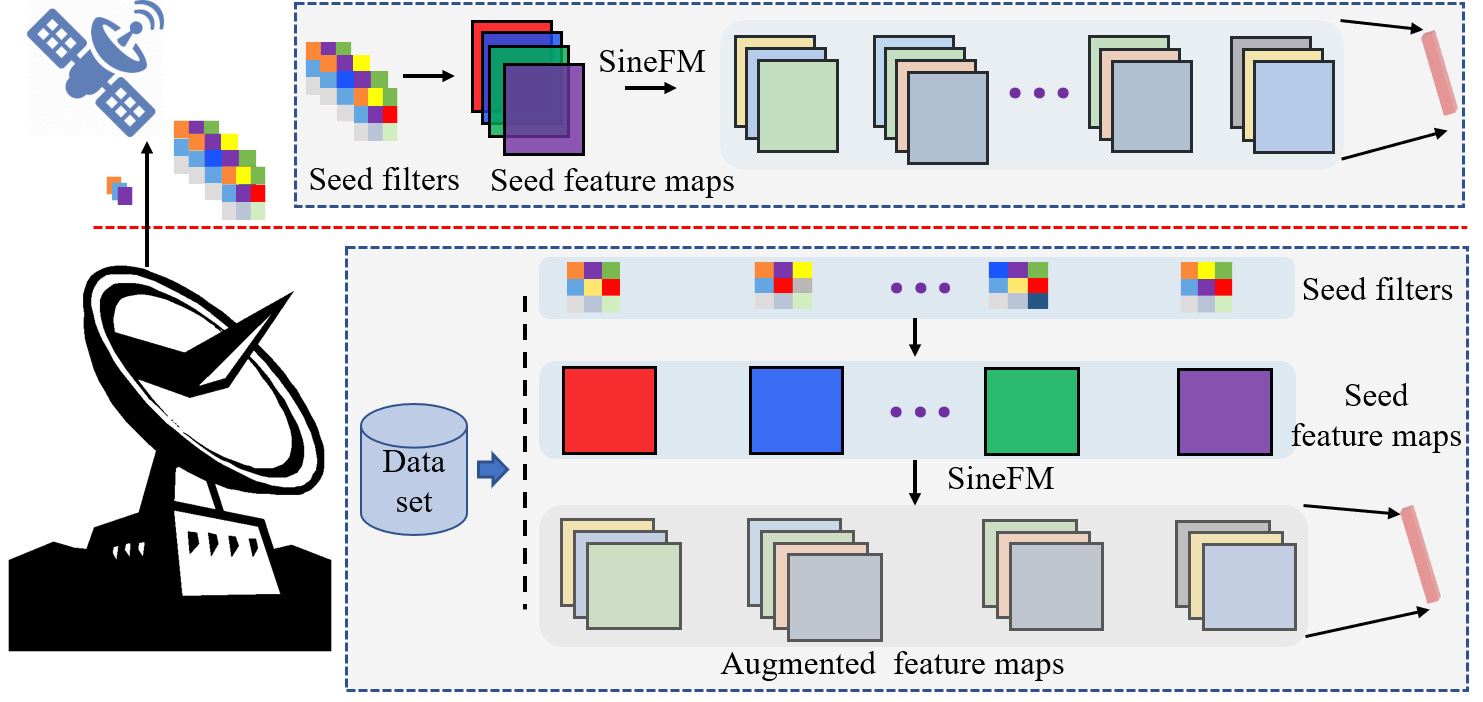}
\caption {Overview of the proposed framework. We first design the ground station server-assisted training CNN model method. Then, we design a SineFN algorithm. Each layer of CNN first learns the weights of one filter (i.e., the seed filter). Then, a feature map (i.e., seed feature map) is generated based on the seed filter and the input. Finally, based on the seed feature map, the SineFM generates other required feature maps. This makes the ground station server only need to send the seed filter and a small number of seeds to the LEO satellites. Subsequently, the LEO satellite generates the well-trained CNN based on the seed filters, seeds, and SineFM.}
\label{fig:framework}
\end{figure*}

\section{Background and Related Work} \label{RelatedWork}
We briefly prior work on convolutional neural network models and server-assisted training and inference, from which our work draws inspiration.

\subsection{Convolution Neural Network Models}
Convolutional neural network (CNN) models~\cite{Alex@ImageNet, Karen@Very, resnet, Huang@Densely} have achieved overwhelming success in fields such as computer vision, natural processing, and speech recognition, and using them to provide high-quality services has become mainstream.
For example, AlexNet~\cite{Alex@ImageNet} won the 2012 ImageNet large-scale vision challenge and greatly improved the accuracy.
VGG~\cite{Karen@Very} proved that increasing the depth can affect the performance of the CNN to a certain extent.
ResNet~\cite{resnet} extends the depth of the CNN to 152 layers by introducing the residual module and significantly improves its performance on multiple vision tasks.
The excellent performance of the above CNN models is largely attributed to the use of convolution operations to extract features.
However, the convolution operation extracts features by sliding filters over the input data, which incurs a large number of FLOPs.
To this end, many alternatives to convolution operations have been proposed~\cite{Howard@MobileNets, mobilenetv2, Howard@Searching, han2020ghostnet, Zhang@ShuffleNet, Ma@ShuffleNet}.  
For example, MobileNets~\cite{Howard@MobileNets, mobilenetv2, Howard@Searching} utilized the depthwise and pointwise convolutions to construct a unit to approximate the standard convolution. 
ShuffleNet~\cite{Zhang@ShuffleNet, Ma@ShuffleNet} further explored a channel shuffling operation to ensure model performance while reducing FLOPs.

Different from the above methods, instead of using filters to slide the input data to obtain feature maps, we use nonlinear transformation to generate feature maps, which can significantly reduce the number of FLOPs for running CNN models.

\subsection{Server-assisted Training and Inference}
Sufficient computing resources are one of the important supports for training high-performance CNNs, and a large number of server-assisted CNN training researches have appeared in recent years~\cite{Ding@A, Liu@A, Kang@Neurosurgeon, Li@JALAD, Laskaridis@SPINN, Teerapittayanon@Distributed, Ding@Cognitive}.
For example, Neurosurgeon~\cite{Kang@Neurosurgeon} divides a CNN model into a head and a tail to run on the device and cloud server respectively.
SPINN~\cite{Laskaridis@SPINN} trains and infers CNN models through the collaboration of devices and the cloud server, and uses compression and quantization techniques to reduce the number of model parameter exchange between the devices and the cloud server.
However, the above methods are highly dependent on the network conditions between the server and the device.
When network conditions are unstable or disconnected, it can greatly degrade the user experience.
To solve this problem, MonoCNN~\cite{Ding@Towards} and TFormer~\cite{Lu@TFormer} are proposed, both of which first train the CNN model on the cloud server, and then send the CNN model to the device for deployment and subsequent update.
For example, MonoCNN is the first to introduce seed filters and random seeds to generate other filters to reduce the number of model parameters sent from the server to the device. 

Different from the above methods, our method uses random seeds to store and reproduce hyperparameters of nonlinear transformation, so that the ground station server only needs to send a small number of parameters to the LEO satellites.
In addition, compared with MonoCNN, using nonlinear transformation to generate feature maps greatly reduces the FLOPs required to run CNN models.

\section{Design of the proposed approach} \label{Proposed Approach}

\subsection{Overview}
Fig.~\ref{fig:framework} illustrates the architecture of the proposed framework. 
Our goal is to provide a framework suitable for ground station server-assisted LEO satellite deployment and update CNN models to provide high-performance remote sensing image processing services.
To achieve this goal, as shown, the proposed framework first trains a CNN on the ground station server, and then sends the well-trained CNN to the LEO satellite to provide remote sensing services.

In the proposed framework, we first propose a cheap feature map generation algorithm, SineFM.
Then, we propose to save and reproduce the hyperparameters of the nonlinear transformation with random seeds to reduce the number of parameters transmitted from the ground station server to LEO satellites.
We describe the proposed framework in detail as follows: (i) the ground station server-assisted method in Section~\ref{ref:groundstationserver-assisted}, (ii) the SineFM algorithm in Section~\ref{ref:PolyFM}.

\subsection{Design of ground station server-assisted method} \label{ref:groundstationserver-assisted}
We use the ground station server to train the CNN first, and then send the well-trained CNN to the LEO satellites based on the following three considerations.
(i) Due to the limitation of size and weight, it is difficult for LEO satellites to provide sufficient computing resources for training high-performance CNN models. However, the ground station server has sufficient computing resources.
(ii) LEO satellite-ground bandwidth is limited and frequency band resources are scarce. However, LEO satellite-ground collaborative training or inference CNN models lead to a large amount of satellite-ground data transmission.
(iii) Lightweight CNN model technology is developing rapidly. It is mainstream to train lightweight CNN models on the ground station server and then send them to LEO satellites for inference to provide fast response services.

% \begin {figure}[t]
% \centering
% \includegraphics[width=0.98\linewidth]{figs/latefanout_pipeline.pdf}
% \caption{\textcolor{red}{I'll modify this figure later.}\label{fig:latefanout_pipeline}}
% \end{figure}

\subsection{Design of SineFM} \label{ref:PolyFM}
Convolution operations require a large number of floating-point operations (FLOPs), which is an important indicator to measure the computing resources required by the CNN model.
Here, we count the floating-point operation of one multiplication and one addition as one FLOP.
At any given convolutional layer, the FLOPs are equal to $C_i\times K^2\times W\times H\times C_o$, where $C_i$ is the number of input channels, $K$ is the size of the convolutional filter, $W$ and $H$ are the height and width of the output feature maps, respectively, and $C_o$ is the number of output channels.
Typically, a high-performance CNN contains hundreds of convolutional layers, resulting in a large number of FLOPs.
This makes CNN models unsuitable to run on resource-constrained LEO satellites.

To this end, we propose SineFM algorithm, whose goal is to replace the use of sliding windows to generate feature maps with nonlinear transformations to reduce the FLOPs required to run CNN models.
Next, we explain the SineFM from four parts: generating feature maps through nonlinear transformation, SineFM-based layer, nonlinear transformation, and theoretical analysis.

\subsubsection{Generating feature maps via nonlinear transformation}\label{ref:feature_generation}
At any given layer, we learn a small number of feature maps $\bm{y}^{(\rm{s})}$ by convolving the inputs with a small number of convolutional filters $\bm{W}^{(\rm{s})}$ (i.e, \emph{seed filters}), 
\begin{align}
    \bm{y}^{(\rm{s})} & = \bm{W}^{(\rm{s})}  \ast \bm{x}
\end{align}
where $\bm{x}$ are the inputs. 
Then, we generate the remaining feature maps $\bm{y}^{(\rm{g})}$ by applying nonlinear transformations $\phi(\cdot)$ of the seed feature maps $\bm{y}^{(\rm{s})}$. 
For simplicity, we use monomial functions as the choice of $\phi(\cdot)$ to illustrate the concept. 
Thereafter, the forward propagation for this layer can be mathematically formulated as 
\begin{align}
    y^{(\rm{g})}_i &= \phi\big(y^{(\rm{s})}_i\big) = y^{(\rm{s}){\circ \beta_i} }_i \\
    \tilde{y}^{(\rm{g})}_i &= \frac{y^{(\rm{g})}_i-\frac{1}{n}\sum_i y^{(\rm{g})}_i}{\left(\sum_i\left(y^{(\rm{g})}_i-\frac{1}{n}\sum_i y^{(\rm{g})}_i\right)^2\right)^{\frac{1}{2}}}
\end{align}
where $\beta_i$ is the exponent of the monomial function applied to the $i^{\rm{th}}$ channel of the seed feature maps. 
We also normalize the response maps (i.e., zero mean and unit variance) to prevent the responses from vanishing or exploding, and the normalized response map is now referred to as $\tilde{\bm{y}}^{(\rm{g})}$.
On the other hand, backpropagation through such a nonlinear transformation of the seed feature maps w.r.t a loss function $\ell$ (e.g., cross-entropy) requires the computation of ${\partial \ell}/{\partial \bm{y}^{(\rm{s})}}$, where
\begin{align}
\frac{\partial \ell}{\partial y^{(\rm{s})}_i} &= \frac{\partial \ell}{\partial \tilde{y}^{(\rm{g})}_i}\frac{\partial \tilde{y}^{(\rm{g})}_i}{\partial y^{(\rm{g})}_i}\frac{\partial y^{(\rm{g})}_i}{\partial y^{(\rm{s})}_i} \\
\frac{\partial \tilde{y}^{(\rm{g})}_i}{\partial y^{(\rm{g})}_i} &= \frac{1-\frac{1}{n}}{\left(\sum_i\left(y^{(\rm{g})}_i-\frac{1}{n}\sum_i y^{(\rm{g})}_i\right)^2\right)^{\frac{1}{2}}} \nonumber \\
&- \frac{\left(1-\frac{1}{n}\right)\left(y^{(\rm{g})}_i-\frac{1}{n}\sum_js[j] \right)}{\left(\sum_i\left(y^{(\rm{g})}_i-\frac{1}{n}\sum_i y^{(\rm{g})}_i\right)^2\right)^{\frac{3}{2}}} \\
\frac{\partial y^{(\rm{g})}_i}{\partial y^{(\rm{s})}_i} &= \beta_i y^{(\rm{s}){\circ \left(\beta_i-1\right)}}_i
\end{align}

\subsubsection{SineFM-based layer}
The core idea of the SineFM\footnote{In this paper, we omit the bias terms of convolutional filters for brevity.} is to restrict the CNN network to learn only one (or a few) feature maps at each layer, and through nonlinear transformations, we can generate or augment as many feature maps as needed at each layer. 
The gist is that there are no parameters associated with the augmented features that need to be updated or learned since they are entirely generated by nonlinear transformations and are no longer updated with the back-propagation procedure. 
The hyperparameters of the nonlinear transformation can be saved and reproduced by using several random seeds, which makes the ground station server only needs to send a small number of seed filters and random seeds to the LEO satellites to complete the deployment and update of the model.
Therefore, SineFM is suitable for ground station server to assist LEO satellites deployment and update CNN models.

As shown in Fig.~\ref{fig:framework}, our SineFM-based layer starts with just one (or a few) seed feature map $\bm{y}^{(\rm{s})}$ learned by the seed filters $\bm{W}^{(\rm{s})}$. 
If we desire $m$ feature maps in total for one layer, the $m-1$ feature maps ${\bm{y}}^{(\rm{g})}$ are nonlearnable and are the nonlinear transformation of the seed feature map $\bm{y}^{(\rm{s})}$. 
In our implementation, we include the seed filter as part of the filter bank in the forward propagation, i.e., $[\bm{y}^{(\rm{s})}, \tilde{\bm{y}}^{(\rm{g})}]$. However, it could be excluded if chosen. 
The inputs $\bm{x}$ are initially processed by these seed filters and become $m$ response maps, which are then passed through an activation gate, such as ReLU $\sigma_{\mathrm{relu}}$, and become $m$ feature maps. 
The $m$ feature maps are linearly combined using $m$ learnable weights, which can be operationalized through a $1\times1$ convolution layer.

\begin{align}
    \bm{y} = \sum_{i=1}^m \alpha_i \sigma_{\mathrm{relu}}\left(\Big[\bm{W}^{(\rm{s})}\ast\bm{x}, \phi\left(\bm{W}^{(\rm{s})}\ast\bm{x}\right)\Big]\right)
\end{align}
\noindent where $\alpha$ represents the $1\times1$ convolution weights, $\alpha(\cdot)$ is an activation, and $\phi(\cdot)$ is the nonlinear transformation function. The PyTorch-like pseudocode of SineFM-based layer is demonstrated in Algorithm~\ref{alg:code}.

In addition, through a lot of experiments, we find that replacing the convolutional layer with the SineFM-based layer can improve the performance of the model.
This is because the rule of the nonlinear transformation function acts to regularize the model, thereby improving the performance of the model.

%======================================================
\begin{algorithm}[t]
\caption{SineFM-based layer: PyTorch-like Pseudocode\label{alg:code}}
\definecolor{codeblue}{rgb}{0.25,0.5,0.5}
\definecolor{codekw}{rgb}{0.85, 0.18, 0.50}
\begin{lstlisting}[language=python, mathescape]
# func: non-linear transformation function
# k: fan-out ratio 
# C_i, C_s, C_o: No. of input/seed/output features
class OurConv2d(nn.Conv2d):
    def __init__(self, C_in, C_out, C_s, func, **kwargs):
        super(OurConv2d, self).__init__(
            C_in, C_out, **kwargs)
        self.weight = None  # ensure non-learnable
        # No. of generated features
        C_g = C_s * (ceil(C_o/C_s) - 1)
        # seed filters for generating seed feasures
        self.conv = nn.Conv2d(C_i, C_s, **kwargs)
        self.linear = nn.Conv2d(k*C_s, C_o, 
            kernel_size=1, **kwargs)
        
    def forward(self, x):
        y_s = self.conv(x) # seed features
        y_g = func(y_s, k) # generate k*C_s features
        y_g = self.linear(normalize(y_g)) # linear combination
        return torch.cat([y_s, y_g])
\end{lstlisting}
\end{algorithm}
%======================================================

\subsubsection{Design of nonlinear transformation function} \label{sec:feature_generation}
\begin{figure}[ht]
    \begin{subfigure}[b]{0.235\textwidth}
    \centering
    \includegraphics[trim={0, 0, 0, 0}, clip, width=\textwidth]{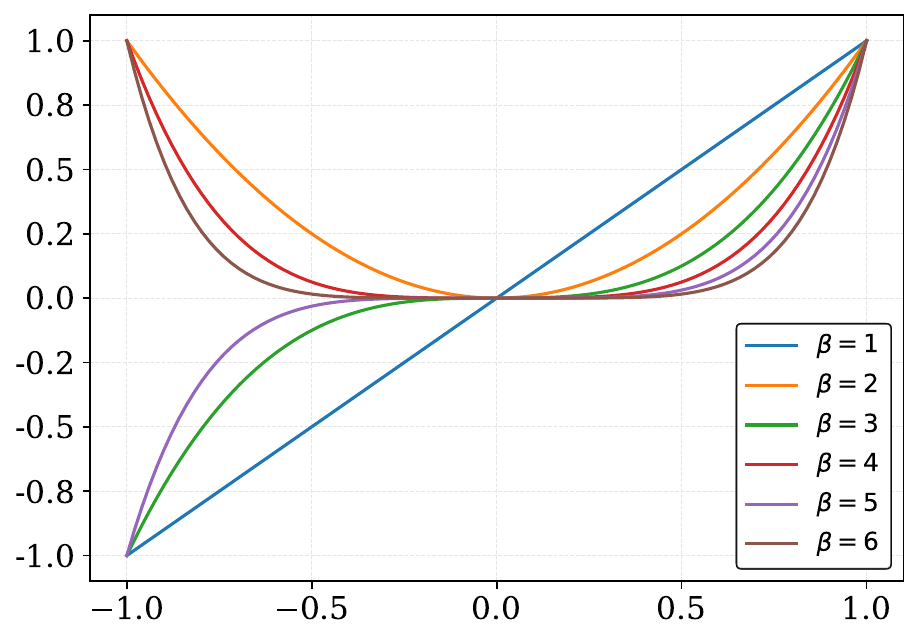}
    \caption{Monomial\label{fig:abl_monomial}}
    \end{subfigure}\hfill
    \begin{subfigure}[b]{0.235\textwidth}
    \centering
    \includegraphics[trim={0, 0, 0, 0}, clip, width=\textwidth]{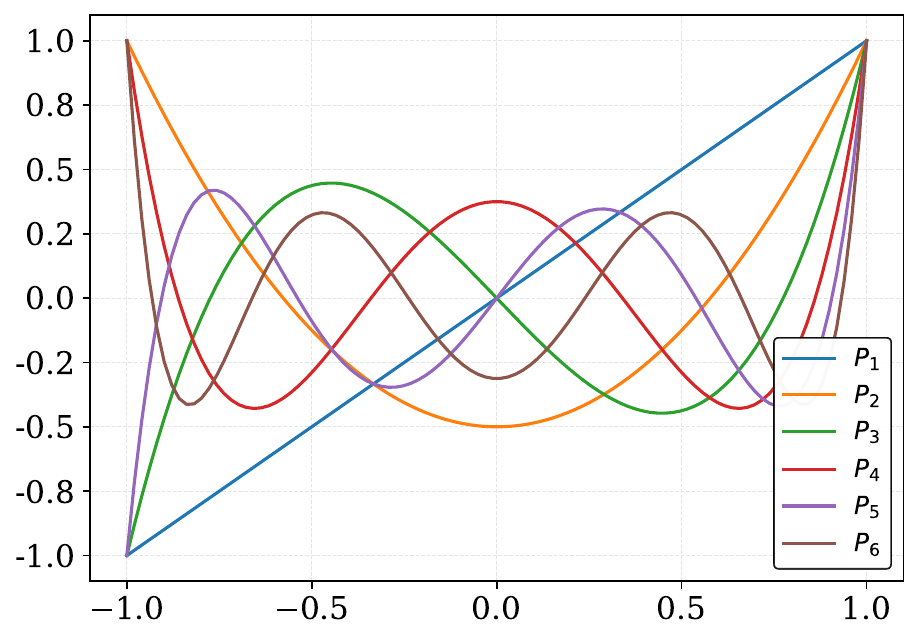}
    \caption{Polynomial (Legendre) \label{fig:abl_polynomial_legendre}}
    \end{subfigure}\\
    \begin{subfigure}[b]{0.235\textwidth}
    \centering
    \includegraphics[trim={0, 0, 0, 0}, clip, width=\textwidth]{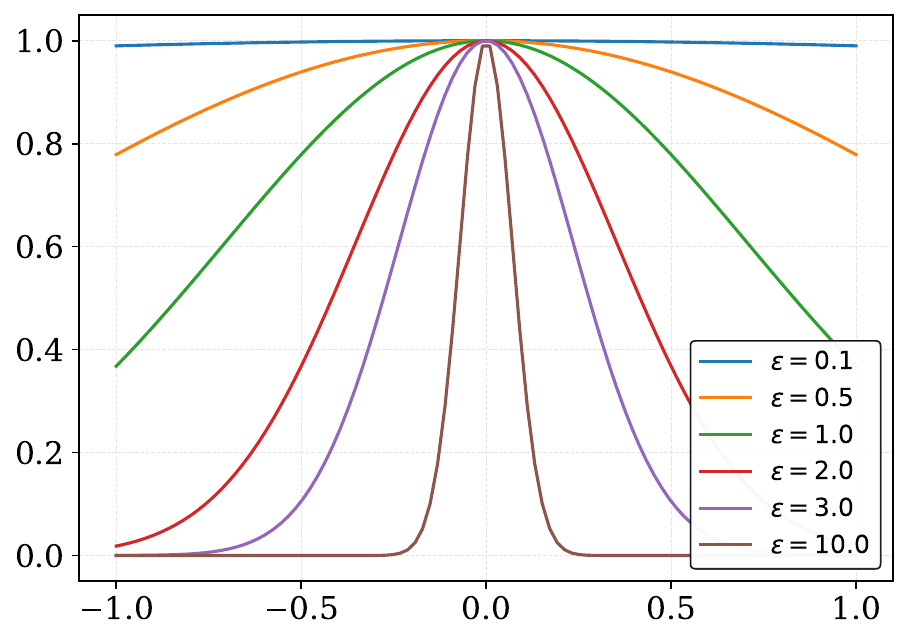}
    \caption{RBF (Gaussian)\label{fig:abl_rbf_gaussian}}
    \end{subfigure} \hfill
    \begin{subfigure}[b]{0.235\textwidth}
    \centering
    \includegraphics[trim={0, 0, 0, 0}, clip, width=\textwidth]{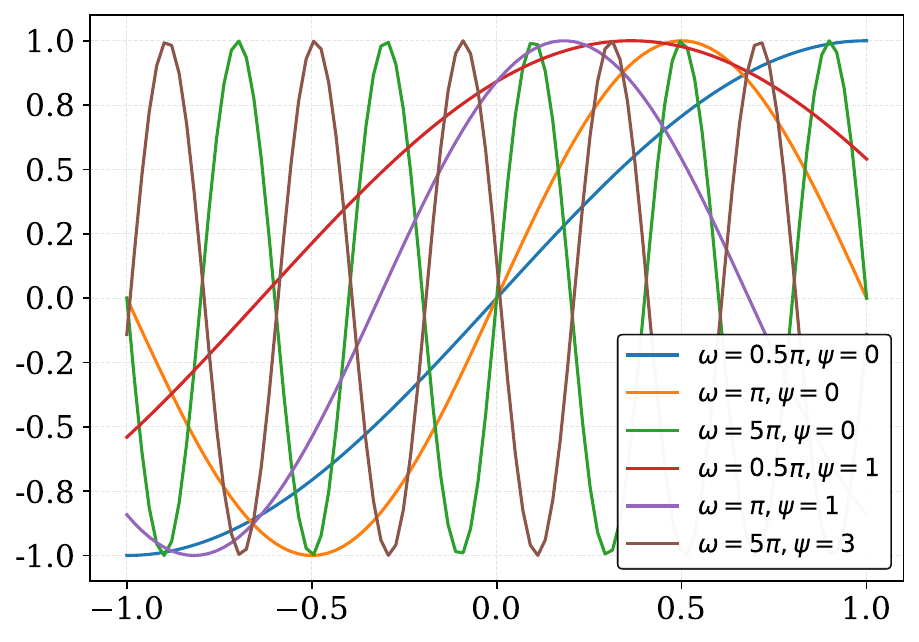}
    \caption{Sinusoidal\label{fig:abl_sine}}
    \end{subfigure} 
    \caption{Visualization of different types of nonlinear transformations. \label{fig:abl_generation_func}}
\end{figure}

There are many nonlinear functions that can be used to generate feature maps.
In this work, we primarily consider monomials, polynomials, radial basis functions, and sinusoidal functions given their well-studied theoretical properties and prevalence in the literature. 
A summary and visualizations of these four types of non-linear transformation functions are provided in Table~\ref{tab:feat_func} and Fig.~\ref{fig:abl_generation_func}, respectively. 
In this work, we empirically determine the choice of feature-generating function through ablation analysis. 
For polynomial and radial basis function families, we have multiple candidates, we first identify the best choice within their respective families. 

\begin{figure*}[ht]
    \begin{subfigure}[b]{0.325\textwidth}
    \centering
    \includegraphics[trim={0, 0, 0, 0}, clip, width=\textwidth]{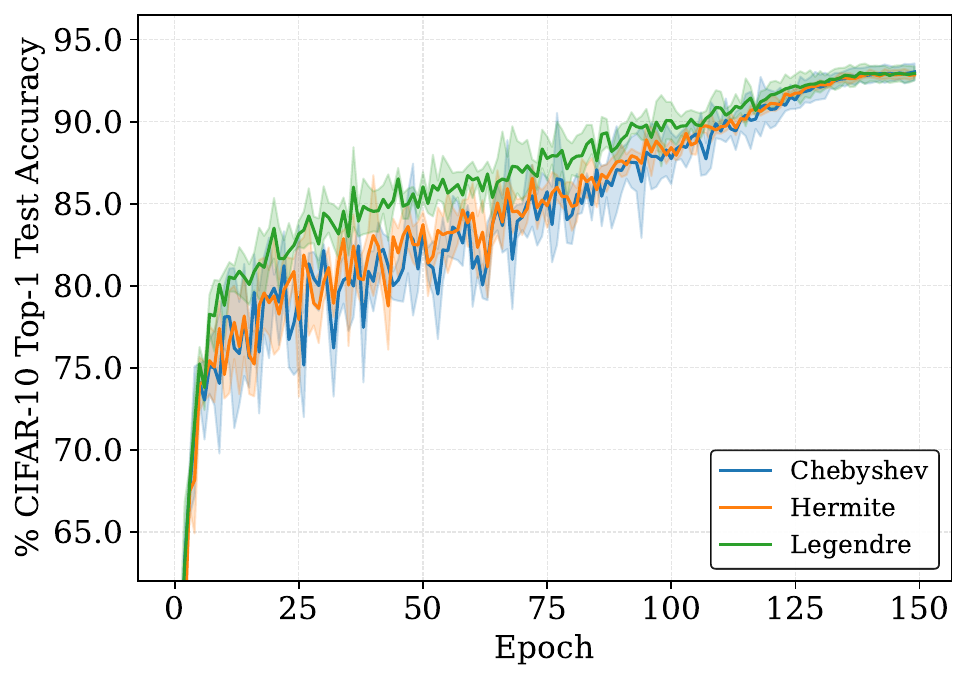}
    \caption{Polynomial function\label{fig:abl_poly}}
    \end{subfigure}\hfill
    \begin{subfigure}[b]{0.325\textwidth}
    \centering
    \includegraphics[trim={0, 0, 0, 0}, clip, width=\textwidth]{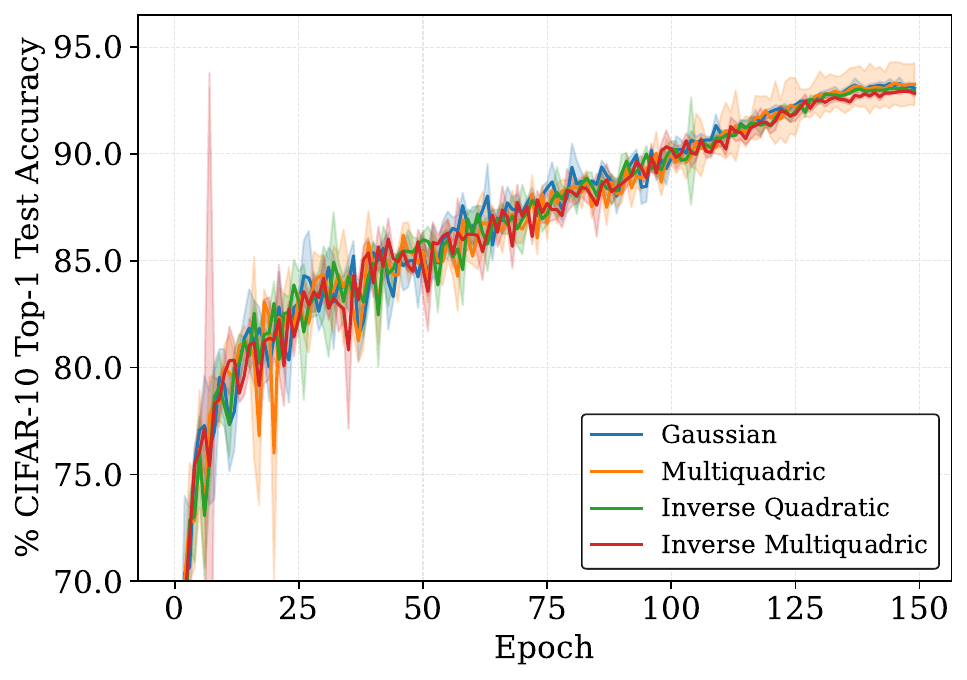}
    \caption{Radial basis function\label{fig:abl_rbf}}
    \end{subfigure}\hfill
    \begin{subfigure}[b]{0.325\textwidth}
    \centering
    \includegraphics[trim={0, 0, 0, 0}, clip, width=\textwidth]{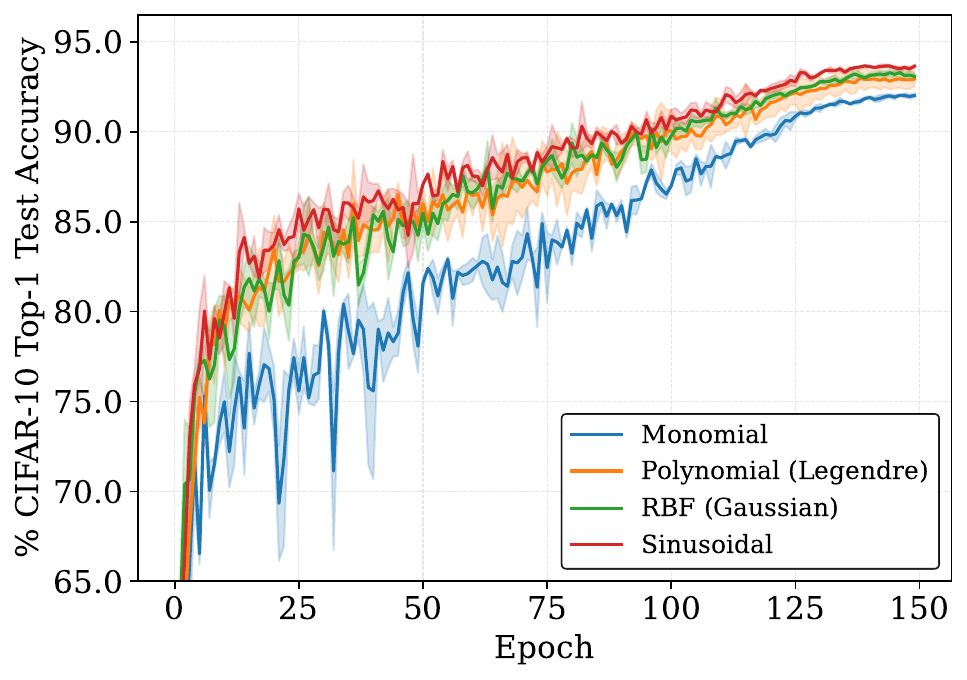}
    \caption{Overall comparison\label{fig:abl_func_c10}}
    \end{subfigure}
    \caption{Comparison of different choices (a) polynomial and (b) radial basis functions for generating features from seed features. (c) Overall comparison among different feature-generating functions. \label{fig:abl_feature_func}}
\end{figure*}

Figs.~\ref{fig:abl_poly} and \ref{fig:abl_rbf} depict the results. Evidently, we observe that (i) Legendre outperforms other alternative polynomial functions and (ii) all radial basis functions perform similarly with Gaussian being slightly more stable. 
Accordingly, we use the Legendre polynomial and Gaussian radial basis function and provide an overall comparison among the four types of nonlinear transformations considered in Fig.~\ref{fig:abl_func_c10}. 
We observe that the sinusoidal function outperforms other candidate functions. Therefore, we use the sinusoidal function to generate feature maps for the main results presented in this work. 

\begin{table}[t]
\centering
\caption{Overview of nonlinear transformation functions explored in this work for generating feature maps. \label{tab:feat_func}}
\resizebox{.485\textwidth}{!}{%
\begin{tabular}{@{}lll@{\hspace{2mm}}}
\toprule
 & Transformation & Formulation \\ \midrule
 & Monomial & $\mathrm{sign}(x)|x|^{\beta}$ \\ \midrule
\parbox[t]{2mm}{\multirow{3}{*}{\rotatebox[origin=c]{90}{\scriptsize Polynomials}}} & Chebyshev & $T_n(x)=\begin{cases}
			cos(n~arccos~x\big), & \text{for $|x|\leq1$}\\
                cosh(n~arcosh~x), & \text{for $x\geq1$}\\
                (-1)^ncosh\big(n~arcosh(-x)\big), & \text{for $x\leq1$}\\
		 \end{cases}$ \\
 & Hermite & $H_n(x)=\left(-1\right)^ne^{x^2}\frac{d^n}{dx^n}e^{-x^2}$ \\
 & Legendre & $P_n(x)=\frac{1}{2^nn!}\frac{d^n}{dx^n}\left(x^2-1\right)^n$ \\ \midrule
\parbox[t]{2mm}{\multirow{5}{*}{\rotatebox[origin=c]{90}{\scriptsize \begin{tabular}[c]{@{}l@{}}Radial basis\\functions\end{tabular}}}} & Gaussian & $e^{-(\epsilon x)^2}$ \\
 & Multiquadratic & $\sqrt{1+(\epsilon x)^2}$ \\
 & Inverse Quadratic & $\frac{1}{1+(\epsilon x)^2}$ \\
 & \begin{tabular}[c]{@{}l@{}}Inverse\\Multiquadratic\end{tabular} & $\frac{1}{\sqrt{1+(\epsilon x)^2}}$ \\ \midrule
 & Sinusoidal & $sin(\omega x+\psi)$ \\ \bottomrule
\end{tabular}%
}
\end{table}

\subsubsection{Theoretical analysis} \label{sec:theory}
In this section, we provide a theoretical analysis of our layer and demonstrate how it can well approximate the standard convolutional layer.

At layer $l$, let $\bm{x}_\pi \in \mathds{R}^{(C \cdot k \cdot k) \times 1}$ be a vectorized single patch from the $C$-channel inputs at location $\pi$, where $k$ is the kernel size of the convolutional filter. Let $\bm{w} \in \mathds{R}^{(C \cdot k \cdot k) \times 1}$ be a vectorized single convolution filter from the convolutional filter tensor $\bm{W} \in \mathds{R}^{C\times k \times k \times m}$, which contains a total of $m$ generated convolutional filters at layer $l$. We drop the layer subscription $l$ for brevity.

In a standard CNN, this patch $\bm{x}_\pi$ is taken as a dot product with the filter $\bm{w}$, followed by the nonlinearity (e.g., ReLU $\sigma_{\mathrm{relu}}$), resulting in a single output feature value $d_\pi$ at the corresponding location $\pi$ on the feature map. Similarly, each value of the output feature map is a direct result of convolving the entire input map $\bm{x}$ with a convolutional filter $\bm{w}$. This microscopic process can be expressed as: 
% -------------------------------------------------------------
\begin{align}
d_\pi = \sigma_{\mathrm{relu}} ( \bm{w}^{\top} \bm{x}_\pi )
\label{eq:d}
\end{align}
% -------------------------------------------------------------

For our proposed layer (we again consider monomials as the choice of $\phi(\cdot)$ feature generation functions for illustration), the seed features are the direct results of convolving the inputs $\bm{x}$ with the seed convolutional filters $\bm{w}^{(\rm{s})}$. Then, we obtain a set of $m$ response maps by applying the monomial transformation to seed features with exponents coefficients of $\beta_1,\beta_2,\ldots,\beta_m$ which are pre-defined and not updated during training. The corresponding output feature map value $\tilde{d}_\pi$ is also a linear combination of the corresponding elements from the $m$ response maps via $1\times1$ convolution with parameters $\alpha_1, \alpha_2, \ldots, \alpha_m$. This process follows:
\begin{align}
\tilde{d}_\pi &= \sigma_{\mathrm{relu}}\underbrace{
\left(\begin{bmatrix}
    ( \bm{w}^{(\rm{s})\top} \bm{x}_{\pi} )^{\circ \beta_1} \\
    ( \bm{w}^{(\rm{s})\top} \bm{x}_{\pi} )^{\circ \beta_2} \\
    \cdots \\
    ( \bm{w}^{(\rm{s})\top} \bm{x}_{\pi} )^{\circ \beta_m}
\end{bmatrix}\right)^\top}_{1\times m} \underbrace{\boldsymbol{\alpha}}_{m\times1} \nonumber\\
&= \sigma_{\mathrm{relu}}\left(\begin{bmatrix}
    \phi_{\beta_1}( \bm{w}^{(\rm{s})})^\top  \phi_{\beta_1}(\bm{x}_{\pi} ) \\
    \phi_{\beta_2}( \bm{w}^{(\rm{s})})^\top  \phi_{\beta_2}(\bm{x}_{\pi} ) \\
    \cdots \\
    \phi_{\beta_m}( \bm{w}^{(\rm{s})})^\top  \phi_{\beta_m}(\bm{x}_{\pi} )
\end{bmatrix}\right)^\top {\boldsymbol{\alpha}} = \bm{c}_{\mathrm{relu}}^\top \boldsymbol{\alpha}
\label{eq:d_late}
\end{align}
where $\phi_{\beta_i}(\cdot)$ is the point-wise monomial expansion with exponent $\beta_i$. 
Comparing $d_\pi$ (Eq~\ref{eq:d}) and $\tilde{d}_\pi$ (Eq~\ref{eq:d_late}), we consider the following two cases (i) when $d_\pi=0$: since $\bm{c}_{\mathrm{relu}} = \sigma_{\mathrm{relu}}\big(\begin{bmatrix}\cdots\end{bmatrix}\big) \geq 0$, there always exists a vector $\boldsymbol{\alpha} \in \mathds{R}^{m\times1}$ such that $\tilde{d}_\pi=d_\pi$. 
However, when (ii) $d_\pi>0$, it is obvious that the approximation does not hold when $\bm{c}_{\mathrm{relu}}=\bm{0}$. 
Thus, under the assumption that $\bm{c}_{\mathrm{relu}}$ is not an all-zero vector, the approximation $\tilde{d}_\pi \approx d_\pi$ will hold.

In summary, SineFM is the core of the proposed framework.
There are three advantages as follows:
First, based on the seed feature map, using nonlinear transformation instead of sliding windows to generate feature maps, which effectively reduces the FLOPs required to run the CNN model.
Second, the hyperparameters of nonlinear transformations are randomly generated and nonlearnable, which allows these parameters to be saved and reproduced with some random seeds, so that only a small number of parameters need to be transmitted from the ground station to the LEO satellite.
Third, the nonlinear transformation plays the role of regularizing the model which can improve its performance.

\iffalse
After the IoT device receives the seed filters and seeds sent by the cloud server, there are two methods for using MonoCNN.
%
The first method is to generate the MonoCNN according to the seed filters, seeds and the FGF when the IoT is idle and store it. 
%
When the MonoCNN model needs to be used, the IoT device can page it into memory to run it in the same way as the standard CNN model.
%
The second method is to dynamically generate the MonoCNN model.
%
That is, when the MonoCNN needs to be used, the IoT device instantly generates the MonoCNN by paging the seed filters, seeds, and the FGF into memory.
%
The second method, which only stores seed filters and seeds on the IoT device, can save memory usage and page-in overhead.
%
However, the price is that there is a certain overhead in generating the MonoCNN model.
%
Practically, since the generation process of MonoCNN has only one multiplication and addition operation, its generation overhead is small.
%
We will test the resource overhead of generating MonoCNN on the IoT device as our future work.
\fi

\section{Experimental Evaluation} \label{Evaluation}
In this section, we first introduce our experimental setup including the datasets, baselines, and evaluation metrics, followed by the implementation details.
Then, we provide an empirical comparison in terms of FLOPs, learnable parameters, and performance on multiple remote sensing datasets.

\subsection{Experimental Setup}
\begin{figure*}[ht]
    \begin{subfigure}[b]{0.18\textwidth}
    \centering
    \includegraphics[height=0.2\textheight]{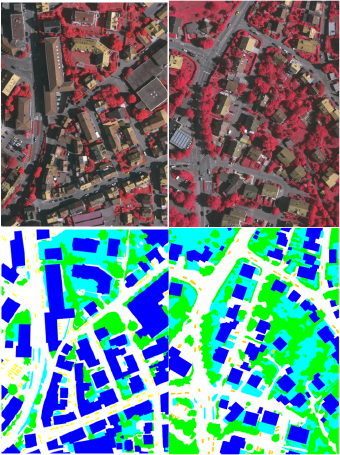}
    \caption{ISPRS Vaihingen\label{fig:vaihingen}}
    \end{subfigure} \hfill
    \begin{subfigure}[b]{0.235\textwidth}
    \centering
    \includegraphics[height=0.2\textheight]{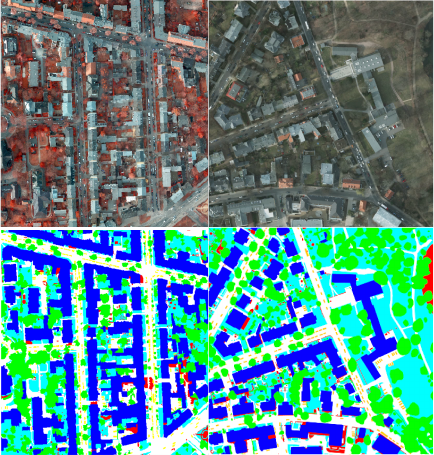}
    \caption{ISPRS Potsdam\label{fig:potsdam}}
    \end{subfigure} \hfill
    \begin{subfigure}[b]{0.22\textwidth}
    \centering
    \includegraphics[height=0.2\textheight]{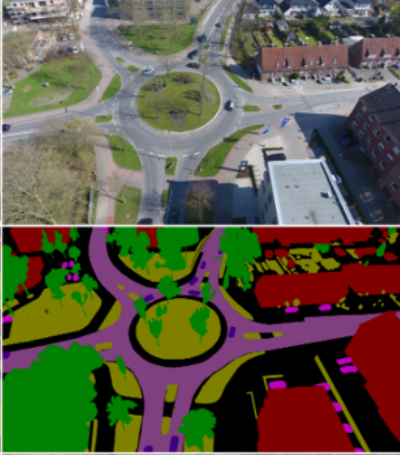}
    \caption{UAVid\label{fig:UAVid}}
    \end{subfigure} \hfill
    \begin{subfigure}[b]{0.27\textwidth}
    \centering
    \includegraphics[height=0.2\textheight]{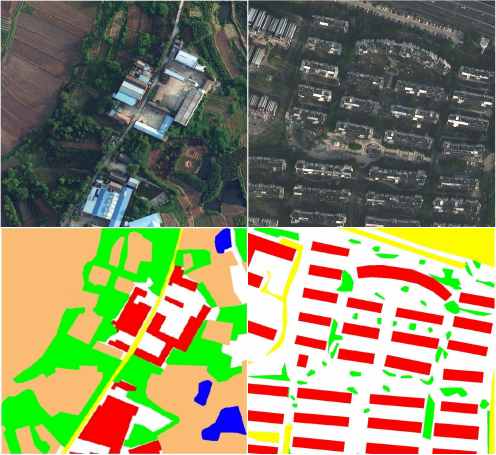}
    \caption{LoveDA\label{fig:LoveDA}}
    \end{subfigure}
    \caption{\textbf{Visualization of Datasets.} The input images are shown in the first row, and the ground truth labels are shown in the second row.\label{fig:dataset}}
\end{figure*}

\noindent\textbf{Datasets.} Four widely-used remote sensing datasets are considered for evaluating the efficacy of the proposed approach. Sample images of these datasets are provided in Fig.~\ref{fig:dataset}.

\emph{ISPRS Vaihingen} \cite{vaihingen} dataset contains 33 high-quality images with topographical information, each with an average resolution of 2494$\times$2064 pixels. These images, referred to as "TOP image tiles," have three color channels (i.e., near-infrared, red, green), as well as a digital surface model and a normalized digital surface model (NDSM) with a ground sampling distance of 9 cm. The dataset includes five categories of objects in the foreground (i.e., impervious surface, building, low vegetation, tree, car) and one category for the background (i.e., clutter). In the experiments conducted, only the TOP image tiles were used, and the DSM and NDSM were disregarded. The images used for testing were identified by the following IDs: 2, 4, 6, 8, 10, 12, 14, 16, 20, 22, 24, 27, 29, 31, 33, 35, and 38.

\emph{ISPRS Potsdam} \cite{potsdam} dataset includes 38 high-resolution aerial images (with a resolution of 5 cm per pixel) that are 6000$\times$6000 pixels in size. These images include the same categories as the Vaihingen dataset. The dataset includes four types of data: red, green, blue, and near-infrared multispectral bands, as well as digital surface models and normalized digital surface models. For testing purposes, we used images ID: 2\_13, 2\_14, 3\_13, 3\_14, 4\_13, 4\_14, 4\_15, 5\_13, 5\_14, 5\_15, 6\_13, 6\_14, 6\_15, and 7\_13, and the remaining 23 images\footnote{We exclude image id 7\_10 due to error annotations.} were used for training. In the experiments, only the red, green, and blue bands were used, and the images were cropped into 1024$\times$1024 pixel sections.

\emph{UAVid} \cite{LYU2020108} is a dataset for semantic segmentation of Unmanned Aerial Vehicles (UAVs) with high resolution. It mainly features urban street scenes with two different sizes of images (3840$\times$2160 and 4096$\times$2160) and 8 different categories. The segmentation of UAVid images presents a challenge due to their high spatial resolution, diverse spatial variations, unclear category definitions, and complex scenes. The dataset consists of 42 sequences, with a total of 420 images. Among these, 200 images are used for training, 70 images are used for validation, and the remaining 150 images are designated for testing and provided by the official source. In our experiments, each image was divided into 8 patches of 1024$\times$1024 pixels after being padded and cropped.

\emph{LoveDA} \cite{wang2021loveda} dataset consists of 5,987 high-quality optical remote sensing images with a resolution of 0.3 meters per pixel for a total size of 1024$\times$1024 pixels. It includes seven types of land cover, namely buildings, roads, water, barren land, forests, agriculture, and backgrounds. Of these images, 2,522 are designated for training, 1,669 for validation, and 1,796 are provided for testing purposes. The dataset covers two types of landscapes, urban and rural, and has been collected from three cities in China: Nanjing, Changzhou, and Wuhan. Due to the multi-scale features of objects, complex backgrounds, and varied distributions of land cover categories, this dataset presents significant challenges.

\vspace{3pt}
\noindent\textbf{Baselines.} We considered a diverse set of representative baselines collected from the literature, including methods that were developed for general computer vision tasks (e.g., GhostNet \cite{han2020ghostnet} and MonoCNN \cite{ding2022towards}), methods that were tailored for efficient semantic segmentation based solely on CNNs (e.g., BiSeNet \cite{yu2018bisenet} and SwiftNet \cite{orvsic2021efficient}) and assisted by attention modules (e.g., DANet \cite{fu2019dual} and MANet \cite{li2021multiattention}) or transformers (e.g., Segmenter \cite{strudel2021segmenter} and SegFormer \cite{xie2021segformer}), and methods that were dedicated to remote sensing, such as UNetFormer \cite{wang2022unetformer}. 

\vspace{3pt}
\noindent\textbf{Evaluation Metrics.}
Our experiments evaluated the performance of all models in two aspects: computational efficiency and prediction accuracy. To measure computational efficiency, we looked at three factors: the number of model parameters ($^{\#}\rm{Params}$), the number of floating-point operations ($^{\#}\rm{FLOPs}$) required, and the inference speed, which was expressed in terms of frames per second (FPS), for processing a single image. To assess the prediction accuracy, we used three metrics: the mean intersection-over-union (mIoU), the overall accuracy (OA), and the mean F1 score (F1). 

\vspace{3pt}
\noindent\textbf{Implementation Details.}
The experiments were carried out using PyTorch with two NVIDIA GTX 2080Ti GPUs. We used AdamW optimizer with an initial learning rate of $6e^{-4}$, which was gradually reduced to zero following the cosine annealing strategy \cite{loshchilov2017sgdr}. We incorporate our method in the UNetFormer framework \cite{wang2022unetformer} as the backbone feature extractor. For GhostNet \cite{han2020ghostnet}, we set the reduction ratio to two; for MonoCNN \cite{ding2022towards}, we followed the default settings provided in the original paper. The settings for other baselines were adopted from \cite{wang2022unetformer}. 

For the UAVid dataset, data augmentation was performed by applying random vertical and horizontal flips, and random brightness changes to input images. The training epoch was set to 40 and the batch size was 8. During testing, additional augmentations such as vertical and horizontal flips were applied using the test-time augmentation (TTA) technique. For the Vaihinge, Potsdam, and LoveDA datasets, the images were randomly cropped into 512$\times$512 patches. The augmentations included random scaling (among [0.5, 0.75, 1.0, 1.25, 1.5]), random vertical and horizontal flips, and random rotations. The training epoch was set to 100 and the batch size was 16. During testing, multi-scale and random flip augmentations were applied.

\iffalse
The codes are available at \hyperlink{...}{[retracted-for-anonymity]}.
\fi

\subsection{Experimental Results}
\vspace{3pt}
\subsubsection{Improvement in Model Efficiency}

\begin{figure}[t]
    \centering
    \begin{subfigure}[b]{0.4\textwidth}
    \centering
    \includegraphics[trim={0, 0, 0, 0}, clip, width=\textwidth]{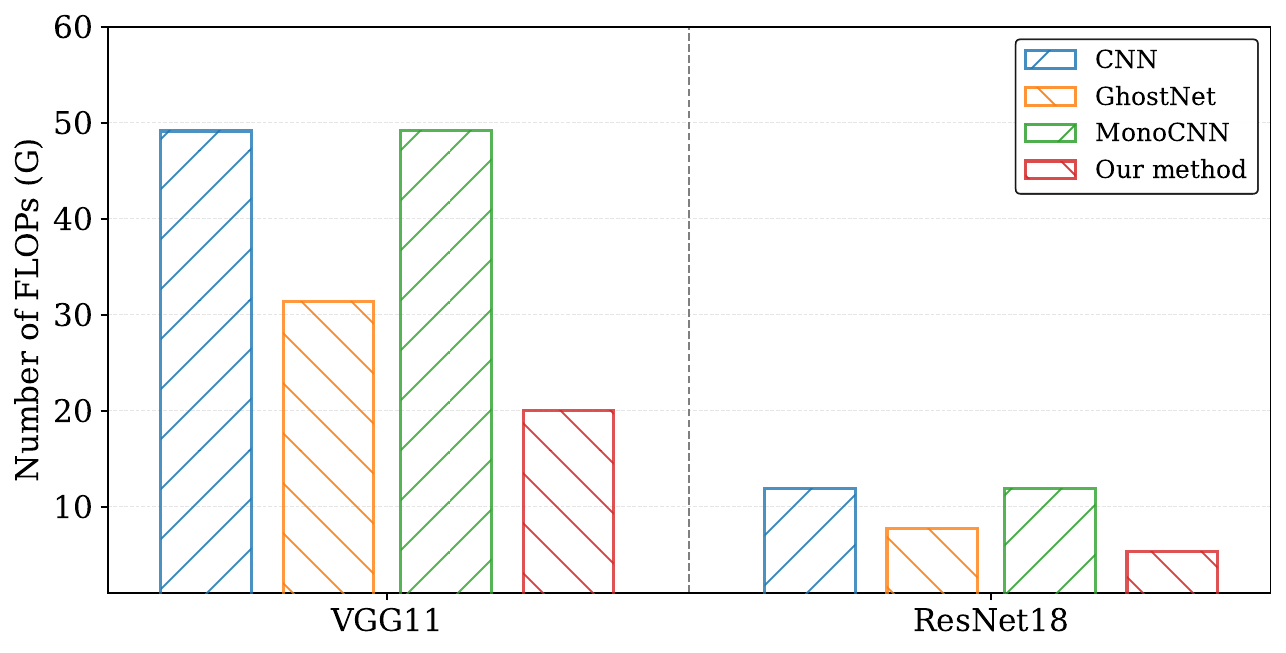}
    \caption{FLOPs (G) \label{fig:compare_flops}}
    \end{subfigure}\\
    \centering
    \begin{subfigure}[b]{0.4\textwidth}
    \centering
    \includegraphics[trim={0, 0, 0, 0}, clip, width=\textwidth]{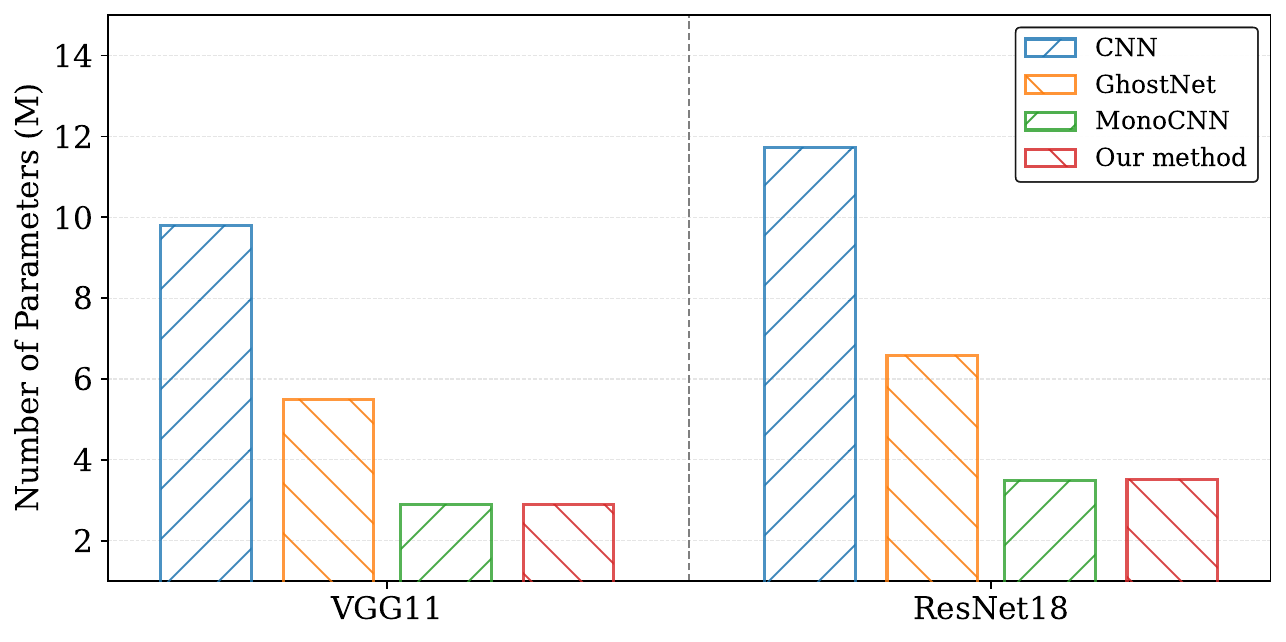}
    \caption{Parameters (M) \label{fig:compare_parameters}}
    \end{subfigure}
    \caption{Comparison of (a) the number of network parameters required to be transmitted and (b) the number of FLOPs required for each prediction. FLOPs are computed w.r.t. input size of $512\times512$.\label{fig:model_efficiency}}
\end{figure}

Our proposed framework achieves the fewest FLOPs and the least amount of model parameter transmission.
As shown in Fig.~\ref{fig:model_efficiency}, we consider two widely used architectures (VGG11 and ResNet18) and compare the FLOPs and the learnable parameters of SineFM-based model for semantic segmentation applications.
Specifically, as shown in Fig.~\ref{fig:compare_flops}, CNN, GhostNet and MonoCNN contain a large number of feature map generation operations using sliding windows, resulting in a large number of FLOPs.
However, our method only requires a sliding operation in each layer of the model to generate a seed feature map, and then uses nonlinear transformations to generate other multiple feature maps based on the seed featur map, avoiding a large number of sliding window operations, thus greatly reducing FLOPs.

As shown in Fig.~\ref{fig:compare_parameters}, 
Since all filter parameters in CNN and GhostNet need to be learned, the ground station server needs to send all filter parameters to LEO satellites, resulting in a large amount of model parameter transmission.
In our proposed framework, only a single-seed filter parameter and a random seed need to be transmitted in each layer.
This is because we can generate a seed feature map based on this seed filter, and then generate other feature maps of this layer based on the seed feature map and nonlinear transformation.
The hyperparameters of the nonlinear transformation are randomly initialized and remain fixed so that these hyperparameters can be saved and reproduced by a random number seed.
Therefore, compared with transmitting all model parameters, our proposed framework can greatly reduce the amount of model parameter transmission.

Additionally, as shown in Fig.~\ref{fig:compare_parameters}, the number of model parameters transmission by MonoCNN is the same as that of our SineFM-based model, because they both need to transmit only a small number of seed filters and seeds. 
The biggest difference between the two is the way to generate feature maps, as shown in Fig.~\ref{fig:compare_flops}, our method reduces more FLOPs.

\subsubsection{Improvement in Model Performance}
Our proposed SineFM-based model consistently outperforms other state-of-the-art models on ISPRS Vaihingen and Potsdam in terms of mean F1, OA, and mIoU. 
As shown in Table~\ref{tab:vaihingen} and Table~\ref{tab:potsdam}, on the ISPRS Vaihingen dataset, compared with BiSeNet, our method achieves a 6.3\% higher mean F1, a 4.4\% higher OA, and a 7.4\% higher mIoU.
On the ISPRS Potsdam dataset, our method achieves a 3.4\% higher mean F1, a 3.6\% higher OA, and a 5.7\% higher mIoU.
This is because the nonlinear transformation acts to regularize the model when generating feature maps using the nonlinear transformation.
Therefore, our method achieves consistently better performance.

\begin{table}[ht]
\centering
\caption{Performance on ISPRS Vaihingen.\label{tab:vaihingen}}
\resizebox{.42\textwidth}{!}{%
\begin{tabular}{@{\hspace{2mm}}lccc@{\hspace{2mm}}}
\toprule
Method & mean F1 ($\uparrow$) & OA ($\uparrow$) & mIoU ($\uparrow$)\\ \midrule
BiSeNet \cite{yu2018bisenet} & 84.3 & 87.1 & 75.8 \\
DANet \cite{fu2019dual} & 79.6 & 88.2 & 69.4 \\
SwiftNet \cite{orvsic2021efficient} & 88.3 & 90.2 & 79.6 \\
ABCNet \cite{li2021abcnet} & 89.5 & 90.7 & 81.3 \\
Segmenter \cite{strudel2021segmenter} & 84.1 & 88.1 & 73.6 \\
UNetFormer \cite{wang2022unetformer} & 90.4 & 91.0 & 82.7 \\
Our method & 90.6 & 91.5 & 83.2 \\ \bottomrule
\end{tabular}%
}
\end{table}

\begin{table}[ht]
\centering
\caption{Performance on ISPRS Potsdam \label{tab:potsdam}}
\resizebox{.42\textwidth}{!}{%
\begin{tabular}{@{\hspace{2mm}}lccc@{\hspace{2mm}}}
\toprule
Method & mean F1 ($\uparrow$) & OA ($\uparrow$) & mIoU ($\uparrow$)\\ \midrule
BiSeNet \cite{yu2018bisenet} & 89.8 & 88.2 & 81.7 \\
DANet \cite{fu2019dual} & 88.9 & 89.1 & 80.3 \\
SwiftNet \cite{orvsic2021efficient} & 91.0 & 89.3 & 83.8 \\
ABCNet \cite{li2021abcnet} & 92.7 & 91.3 & 86.5 \\
Segmenter \cite{strudel2021segmenter} & 89.2 & 88.7 & 80.7 \\
UNetFormer \cite{wang2022unetformer} & 92.8 & 91.3 & 86.8 \\
Our method & 93.2 & 91.8 & 87.4 \\ \bottomrule
\end{tabular}%
}
\end{table}

\begin{figure}[ht]
    \centering
    \begin{subfigure}[b]{0.485\textwidth}
    \centering
    \includegraphics[trim={0, 0, 0, 0}, clip, width=\textwidth]{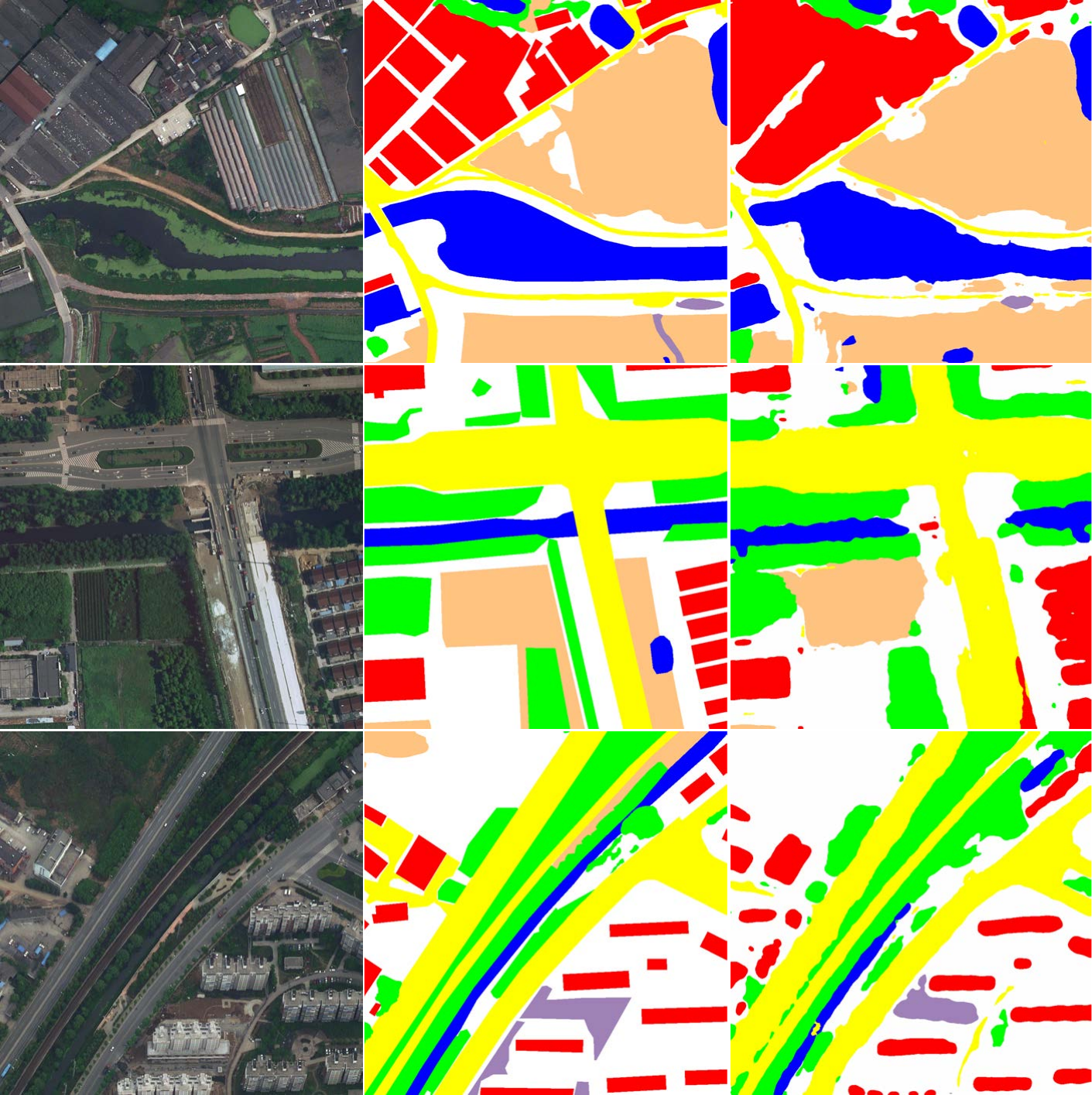}
    \end{subfigure}
    \caption{Qualitative comparison on LoveDA dataset. We show input images, ground truth, and predictions from our method from left to right. \label{fig:loveda_vis_reduced}}
\end{figure}

\vspace{3pt}
\begin{table}[ht]
\centering
\caption{Performance on UAVid. \label{tab:uavid}}
\resizebox{.47\textwidth}{!}{%
\begin{tabular}{@{}llccc@{\hspace{2mm}}}
\toprule
 & Method & $^{\#}\rm{Params}$ ($\downarrow$) & $^{\#}\rm{FLOPs}$ ($\downarrow$) & mIoU ($\uparrow$) \\ \midrule
\parbox[t]{2mm}{\multirow{5}{*}{\rotatebox[origin=c]{90}{\scriptsize\emph{CNN}}}} & BiSeNet \cite{yu2018bisenet} & 12.9M & 51.8G & 63.7 \\
 & DANet \cite{fu2019dual} & 12.6M & 39.6G & 62.8 \\
 & SwiftNet \cite{orvsic2021efficient} & 11.8M & 51.6G & 63.3 \\
 & MANet \cite{li2021multiattention} & 12.0M & 51.7G & 64.8 \\
 & ABCNet \cite{li2021abcnet} & 14.0M & 62.9G & 65.0 \\ \midrule
\parbox[t]{2mm}{\multirow{5}{*}{\rotatebox[origin=c]{90}{\scriptsize\emph{Transformer}}}} & SegFormer & 13.7M & 63.3G & 68.2 \\
 & CoaT \cite{xu2021co} & 11.1M & 105G & 68.0 \\
 & SwinUNet \cite{cao2023swin} & 41.4M & 237G & 68.3 \\
 & UNetFormer \cite{wang2022unetformer} & 11.7M & 46.9G & 70.0 \\
 & Our method & 3.52M & 21.4G & 70.3 \\ \bottomrule
\end{tabular}%
}
\end{table}
\begin{figure}[ht]
    \centering
    \begin{subfigure}[b]{0.485\textwidth}
    \centering
    \includegraphics[trim={0, 0, 0, 0}, clip, width=\textwidth]{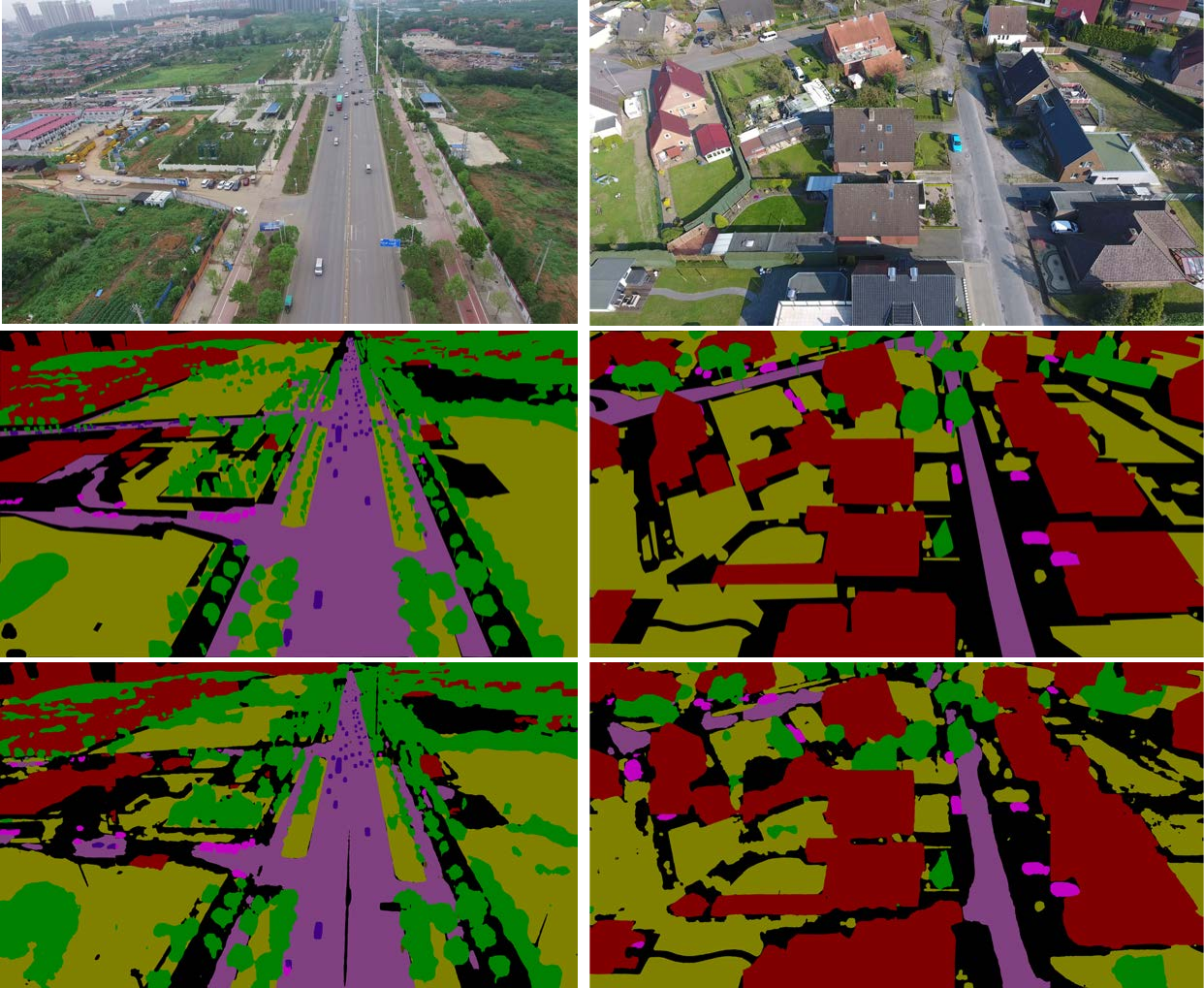}
    \end{subfigure}
    \caption{Qualitative comparison on UAVid dataset. We show input images, ground truth, and predictions from our method from top to bottom. \label{fig:uavid_vis_reduced}}
\end{figure}
 Our proposed SineFM-based model consistently outperforms other state-of-the-art models on UAVid and LoveDA datasets while significantly reducing the number of learnable parameters and FLOPs.
 As shown in Table~\ref{tab:uavid}, compared with the ABCNet, our method improves mIoU by about 5.3\%, but reduces the number of learning parameters by nearly 3$\times$ and FLOPs by 3$\times$.
 Compared with UNetFormer, our method improves mIoU by about 0.3\%, but reduces the number of learnable parameters by nearly 3$\times$ and FLOPs by 2$\times$.
 Akin, as shown in Table~\ref{tab:loveda}, compared with the UNetFormer, our method improves mIoU by 0.4 and achieves higher FPS (+16). 
There are three main reasons for these results.
(i) The feature maps generated by nonlinear transformations are replaced by sliding window-based feature maps, which effectively reduce the FLOPs of running the model.
(ii) The nonlinear transformation parameters are randomly generated and fixed, which effectively reduces the learnable parameters of the model.
(iii) The nonlinear transformation function regularizes the model, thereby improving the performance of the model.
Therefore, our proposed framework is especially suitable for ground station server to assist LEO satellite deployment and update the CNN models to provide high-quality services.
Additionally, we also provide qualitative visualization on both datasets, as shown in Fig.~\ref{fig:uavid_vis_reduced} and Fig.~\ref{fig:loveda_vis_reduced}.

\begin{table}[ht]
\centering
\caption{Performance on LoveDA. \label{tab:loveda}}
\resizebox{.45\textwidth}{!}{%
\begin{tabular}{@{\hspace{2mm}}lccc@{\hspace{2mm}}}
\toprule
Method & $^{\#}\rm{FLOPs}$ ($\downarrow$) & FPS ($\uparrow$) & mIoU ($\uparrow$)\\ \midrule
PSPNet \cite{zhao2017pyramid} & 106G & 52.2 & 48.3 \\
DeepLabV3+ \cite{chen2018encoder}& 95.8G & 53.7 & 47.6 \\
SemanticFPN \cite{kirillov2019panoptic} & 103G & 52.7 & 48.2 \\
SwinUperNet \cite{liu2021swin} & 349G & 19.5 & 50.0 \\
UNetFormer \cite{wang2022unetformer} & 46.9G & 115 & 52.4 \\
Our method & 21.4G & 131 & 52.8 \\ \bottomrule
\end{tabular}%
}
\end{table}

\subsection{Hyperparameter Analysis}

\begin{figure}[ht]
    \begin{subfigure}[b]{0.235\textwidth}
    \centering
    \includegraphics[trim={0, 0, 0, 0}, clip, width=\textwidth]{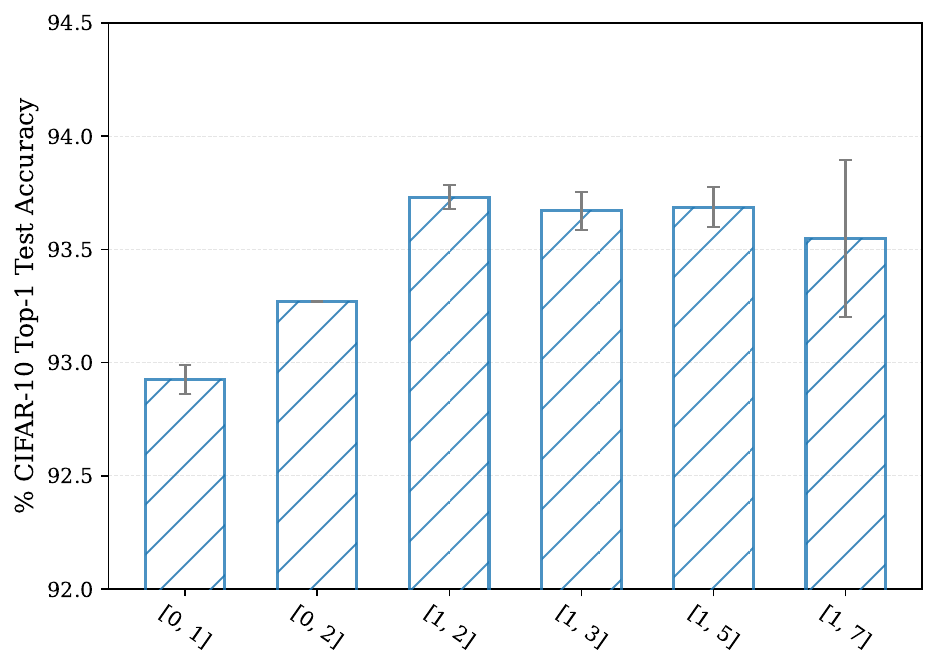}
    \caption{Effect of angular frequency\label{fig:abl_sine_period_range_c10}}
    \end{subfigure}\hfill
    \begin{subfigure}[b]{0.235\textwidth}
    \centering
    \includegraphics[trim={0, 0, 0, 0}, clip, width=\textwidth]{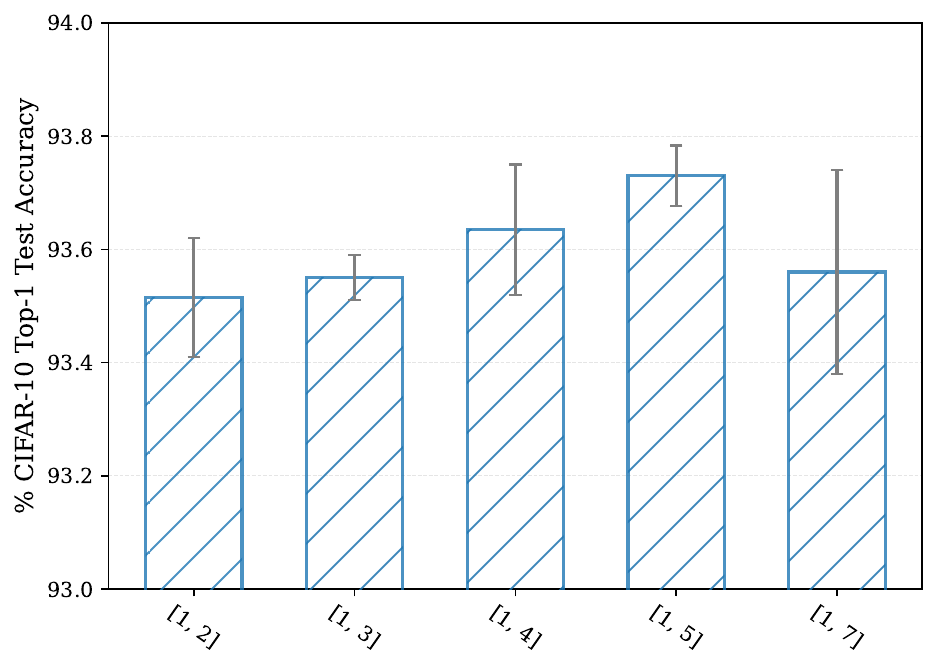}
    \caption{Effect of phase \label{fig:abl_sine_shift_range_c10}}
    \end{subfigure}\\
    \begin{subfigure}[b]{0.235\textwidth}
    \centering
    \includegraphics[trim={0, 0, 0, 0}, clip, width=\textwidth]{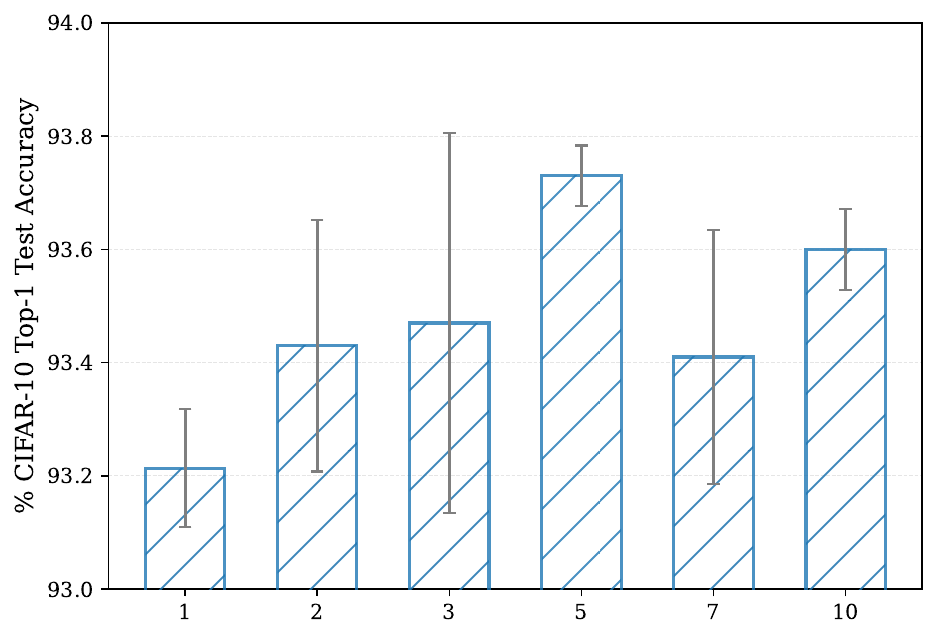}
    \caption{Effect of fan-out ratio\label{fig:abl_sine_fanout_c10}}
    \end{subfigure} \hfill
    \begin{subfigure}[b]{0.235\textwidth}
    \centering
    \includegraphics[trim={0, 0, 0, 0}, clip, width=\textwidth]{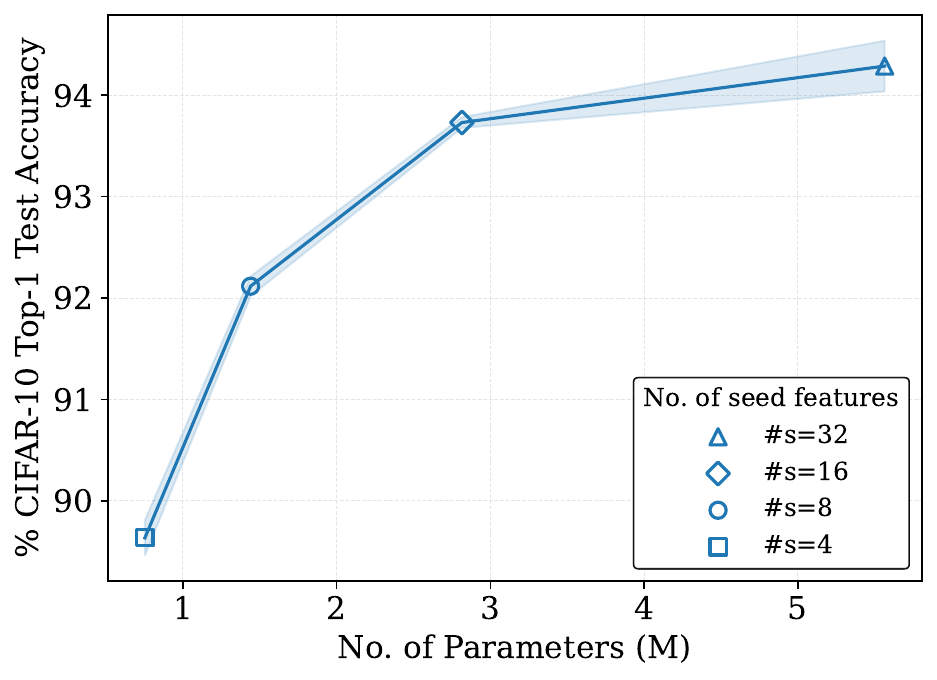}
    \caption{Effect of No. of seed features\label{fig:ablation_seed_features}}
    \end{subfigure} 
    \caption{Ablation studies on various hyperparameters.\label{fig:abl_sine_hyper}}
\end{figure}

The performance of the sinusoidal function for generating features critically depends on the choice of hyperparameters, namely the angular frequency $\omega$ and phase $\psi$ (see Table~\ref{tab:feat_func} for more details). 
For both angular frequency and phase, we randomly sample a $\omega$ and $\psi$ from a uniform distribution of $[a, b]$, where $a$ and $b$ are the lower and upper bounds. 
To understand the effect of $\omega$ and $\psi$, we gradually vary the bounds while keeping all other factors unchanged. 
Figs.~\ref{fig:abl_sine_period_range_c10} and \ref{fig:abl_sine_shift_range_c10} depict the results of $\omega$ and $\psi$, respectively.  
In general, we observe that setting the lower bound $a\geq1$ leads to improved performance. 
In particular, we identify that sampling $\omega$ and $\psi$ from $[1, 2]$ and $[1, 5]$ respectively yields the best result. 

On the other hand, we can create multiple sinusoidal functions with different sets of $\omega$ and $\psi$ at the same time, and we use a hyperparameter (i.e., fan-out ratio) to control this process. 
Ablative analysis has also been carried out to understand the effect of fan-out ratio.
Fig.~\ref{fig:abl_sine_fanout_c10} depicts the results. In general, we observe that more sinusoidal functions simultaneously (i.e., larger fan-out ratio) leads to better performance. Notably, we identify that setting fan-out ratio to five yields the best performance.

Finally, the computational complexity (i.e., the number of parameters and FLOPs) of our method can be controlled via the number of seed filters. 
Fig.~\ref{fig:ablation_seed_features} depicts the trade-off between performance and computational complexity. 
In general, we observe that using 16 seed filters yields the best trade-off in terms of minimizing the number of parameters and maximizing performance. 
\section{Conclusion} \label{Conclusions}
%
\begin{comment}
This paper proposes a ground station server-assisted LEO satellite deployment and update model framework with less model computation and model parameter transmission.
%
The proposed framework first proposes SineFM to generate multiple required feature maps using nonlinear transformation, avoiding the use of sliding windows to generate feature maps to reduce the floating point operations (FLOPs) required to run convolutional neural network (CNN) models.
%
Then, the proposed framework proposes to use several random seeds to save and reproduce the hyperparameters of the nonlinear transformation, which makes the ground station server only needs to send a small number of parameters to the LEO satellite.
%
In addition, the proposed framework analyzes a variety of nonlinear transformation functions in detail and gives a theoretical analysis.
%
Finally, experimental results show that the proposed framework achieves better performance in remote sensing image semantic segmentation applications while reducing the number of FLOPs and parameter transmission.
%
\iffalse
In future work, we plan to deploy the convolutional neural network model generated by the SineFM algorithm on the LEO satellite, and test the network transmission traffic sent to the LEO satellite when the ground station server assists in updating the generated model.
%
\fi
\end{comment}

This paper introduces a novel framework for ground station server-assisted deployment and updating of Low Earth Orbit (LEO) satellites. 
The proposed framework minimizes the computational requirements and reduces the transmission of model parameters by employing the SineFM technique for nonlinear transformation in generating feature maps for Convolutional Neural Network (CNN) models. 
The use of random seeds to save and reproduce the hyperparameters of the nonlinear transformation enables the ground station server to transmit a smaller number of parameters to the LEO satellite.

The proposed framework undergoes a thorough analysis of various nonlinear transformation functions and includes a theoretical analysis to support its implementation. 
Experimental results demonstrate that the proposed framework achieves improved performance in remote sensing image semantic segmentation applications while also reducing the number of floating point operations and parameters transmitted.

In future work, we plan to deploy the CNN model generated by the SineFM algorithm on LEO satellites and evaluate the network transmission traffic sent to the LEO satellite during model updates, assisted by the ground station server. 
This research represents a significant step towards optimizing the deployment and updating processes of LEO satellites and holds great promise for further advancements in the field.


\iffalse
\ifCLASSOPTIONcompsoc
  \section*{Acknowledgments}
\else
  \section*{Acknowledgment}
\fi

The authors would like to thank...
\fi
\section*{Acknowledgements}
This work was supported by the National Natural Science Foundation of China (No. 62202039, 62106097, 62032003), the National Key Research and Development Program of China (No. 2022ZD0118502), and the Fundamental Research Funds for the Central Universities, Sun Yat-sen University (No. 23qnpy94).

% \end{thebibliography}

\bibliography{egbib} 
\bibliographystyle{IEEEtran}

\end{document}